# Performance Enhancement of Distributed Quasi Steady-State Genetic Algorithm

Rahila Patel

M.Tech. IV Sem, CSE, GHRCE, Nagpur (m.s.), India

rahila.patel@gmail.com

M.M.Raghuwanshi

*NYSS College of Engineering and Research, Nagpur (m.s.), India*

m_raghuwanshi@rediffmail.com

Anil N. Jaiswal

*PG Deptt. CSE,GHRCE, Nagpur*

jaiswal_an@yahoo.com

*Urmila Shrawankar*

*PG Deptt. CSE,GHRCE, Nagpur*

urmilas@rediffmail.com

## Abstract

This paper proposes a new scheme for performance enhancement of distributed genetic algorithm (DGA). Initial population is divided in two classes i.e. female and male. Simple distance based clustering is used for cluster formation around females. For reclustering self-adaptive K-means is used, which produces well distributed and well separated clusters. The self-adaptive K-means used for reclustering automatically locates initial position of centroids and number of clusters. Four plans of co-evolution are applied on these clusters independently. Clusters evolve separately. Merging of clusters takes place depending on their performance. For experimentation unimodal and multimodal test functions have been used. Test result show that the new scheme of distribution of population has given better performance.

## 1. Introduction

K-means algorithm is widely used unsupervised partitioning based clustering algorithm. MacQueen proposed k-means in 1967[1]. It uses the Heuristic information to make the search more objective in order that the searching efficiency is improved. Simply speaking it is an algorithm to cluster objects based on attributes/features into K number of group. K is positive integer number. The grouping is done by minimizing the sum of squares of distances between data and the corresponding cluster centroid.

The traditional k-means cluster algorithm has its inherent limitations: 1. Random initialization could lead to different clustering results, even in worst case no result. 2. The algorithm is based on objective function, and usually takes the Gradient method to solve problem. As the Gradient method searched along the direction of energy decreasing (minimization of square error) , that makes the algorithm get into local optimum, and sensitive to isolated points [2]. So the improvement on K-means aims at two aspects: optimization of initialization and improvement on global searching capability.

Distributed GA *(*DGA) is one of the most important representatives of methods based on spatial separation. The basic idea of DGA lies in the partition of the population into several subpopulations, each one of them being processed by a GA, independently from the others. Furthermore, an operator, called migration, produces a chromosome exchange between the sub-populations. Its principal role is to promote genetic diversity, and in essence, to allow the sharing of possible solutions. DGA show two determinant advantages: 1) the preservation of the diversity due to the semi-isolation of the subpopulations, which may stand up to the premature convergence problem, and 2) they may be easily implemented on parallel hardware, obtaining, in this way, substantial improvements on computational time. [3]

As GA implements the idea of evolution, it is natural to expect adaptation to be used not only for finding solutions to a problem, but also tuning the algorithm for the particular problem. Traditional real-coded genetic algorithm (RCGA) has parameters that must be specified before the RCGA is run on a problem. Setting parameter value correctly is a hard task. In general, there are two major forms of setting parameter values: parameter tuning and parameter control [4]. Parameter tuning is the usual approach of finding good parameter values before the run of the GA and then these static values are used during the whole run. Parameter control is different because it changes initial parameter values during the run. Parameter control itself has three variants: deterministic, adaptive and self-adaptive. Deterministic means that parameters are changed according to a rule that uses no feedback from the GA. For adaptive control, feedback does take place and the values are changed in direction or magnitude depending on this feedback. Self-adaptivity is the idea of evolution of the parameters of evolution; the values are encoded into the chromosomes and are also subject to mutation, crossover and selection. The better parameter values will survive because they produce better offspring [5].

In this paper we have suggested self-adaptation for K-means algorithm. It overcomes problems associated with traditional k-means, when used with gendered GA. Different genetic algorithms are proposed, designed and implemented for the single objective as well as for the multiobjective problems. GAS3 [7] (Genetic Algorithm with Species and Sexual Selection) is a distributed Quasi steady state real-coded genetic algorithm. We have applied self-adaptive k-means to GAS3 algorithm for reclustering of population to preserve diversity and to avoid premature convergence.

The organization of this paper is as follows: Section 2 briefly describes the basics of distributed GA; Section 3 comments on improved K-means algorithm; Section 4 applying K-means to GAS3 algorithm; Section 5 gives comparison of results of proposed algorithm GAS3KM over unimodal and multimodal test functions with GAS3 algorithm and finally we draw some conclusions.

## 2. Distributed Genetic Algorithm

Distributed GA (DGA) partitions the population into several subpopulations, each one of them being processed by a GA, independently of the others. Furthermore, a *migration/merging* mechanism produces a chromosome exchange between the subpopulations. DGA attempts to overcome premature convergence by preserving diversity due to the semi-isolation of the subpopulations. Another important advantage is that they may be implemented easily on parallel hardware [6].

Many researchers contributed in the development of distributed genetic algorithms. Brief about few efforts are mentioned over here. One can find detailed study on parallel GA in [8]. Paper [9] presents a genetic algorithm, which is distributed in two novel ways: along genotype and temporal axes. This algorithm divides the candidate solutions into pipelined sets and thus the distribution is in the temporal domain, rather than that in the spatial domain.

Distributed GA is the best choice for implementation of real world problems of distributed domains. Paper [10] presents DGA for energy-efficient resource management in an environment monitoring and hazard detection sensor network. The optimization algorithm is designed based on the Island multi-deme genetic algorithm (GA).

Distributed Genetic Algorithm (DGA) is one of the most promising choices among the optimization methods. In [11] authors describe DGA Frame, a flexible framework for evolutionary computation, written in Java. DGA Frame executes GAs across a range of machines communicating through RMI network technology, allowing the implementation of portable, flexible GAs that use the island model approach.

When the global population is divided into small sub-populations, the ability of these sub-populations to evolve is set back because of their relatively small sizes. The sizes of these subpopulations are dynamically changed according to their performance in [12]. Better sub-populations get more quotas of the total number of individuals, thus get more possibility to produce even better ones. Dynamic rearrangement of the global population can make the process of evolution faster.

## 3. Improved K-means Algorithm

The performance of evolutionary algorithms depends on the characteristics of the population's distribution. Self-Adaptation aims at biasing the distribution towards appropriate regions of the search space by maintaining sufficient diversity among individuals in order to enable further evolvability. Generally, this is achieved by adjusting the setting of control parameters. The ability of an evolutionary algorithm to adapt its search strategy during the optimization process is a key concept in evolutionary computation. Online adaptation of strategy parameters is important, because the best setting of an EA is usually not known a priori for a given task and a constant search strategy is usually not optimal during the evolutionary process. To support this we propose self-adaptive k-means algorithm, which automatically detects initial position of cluster centre and number of clusters.

This algorithm is based on an adaptive approach where dataset is divided in two classes, females and males, based on certain criteria. No need to provide parameters: initial positions of cluster center and number of clusters. It automatically selects females as cluster centers and cluster formation takes place around these females. So number of clusters is equal to number of females. To overcome problem associated with random choice of initial cluster centers, fertile females are selected as initial cluster centers. Again if data set is large then it is very difficult to guess proper number of clusters a priori. This problem is also tackled by providing number of females as number of clusters.

The self-adaptive K-means we have used is given as:
1. Select only those individuals of population, which are marked as female, using Sex Determination Method (SDM), as cluster centres.
2. Assign males to the nearest female.
3. For every cluster
   3.1 Find mean of males and female i.e. mean of cluster.
   3.2 Update female with the mean of cluster.

4. Repeat steps 2 and 3 until females stop moving.

Each cluster consists of one female as cluster center and number of males, which are near to that female. The population P of size N is divided into two classes, female (maximum number of female= k) and male (number of males=(n=N-k)). This algorithm aims at minimizing an *objective function*; in this case a squared error function. The objective function

$$J = \sum_{j=1}^{k}\sum_{i=1}^{n}\left\|x_i^{(j)} - c_j\right\|^2,$$

where $\left\|x_i^{(j)} - c_j\right\|^2$ is a chosen distance measure between a male $x_i^{(j)}$ and the female $c_j$, is an indicator of the distance of the *n* males from their respective females. One can find details of k-means algorithm in [13], [14] and its different usage and various improvements in K-means clustering algorithm in [15]-[21].

## 4. Applying K-means to GAS3 Algorithm

GAS3 is a distributed quasi steady-state genetic algorithm with species and sexual selection scheme. Each of the iteration generates offspring in each species and updates sub-population. There is a generation gap but not as wide as in generational GAs that makes GAS3 quasi steady state GA.

GAS3 uses sex determination method (SDM) to determine the sex (male or female) of members in population. One can find different selection strategy in [22][23][24]. The method is outline as follows:

Let N is population size. Let R refers to the parameter responsible for setting selection pressure (S) in SMD.
1. S=N/R
2. For i= 1 to S do
For j=1 to N do
 a. Take $j^{th}$ member and select four members randomly form population.
 b. Perform recombination operation to produce offspring.
 c. If offspring is better than $j^{th}$ member then it will replace $j^{th}$ member in population and also increments performance count of $j^{th}$ member by one.
3. For i=1 to N do
If (fertility count of $i^{th}$ member > average fertility count) then
$i^{th}$ Member is female
Else
$i^{th}$ Member is male

High value of S (i.e. R is low) gives more chances to an individual to produce offspring. SDM is deterministic in nature and each female represent an elite solution in the population.

Each female member is considered as a niche in the population and the species formation takes place around these niches. Species formation is based on Euclidean distance between female and male members. Each species contains one female member and zero or more male members. All species gets equal chances to produce offspring using Parent Centric Crossover Operator (PCCO) [25] or mutation operator. The sexual selection strategy selects female and required number of male members, from species to perform recombination operation. After a certain generation the performance of the species is evaluated. If species is not performing well, then the merging to the nearby species takes place. The parent centric self-adaptive multi-parent recombination operators are used to explore the search space. The species formation, merging and sexual selection technique helps to exploit a search space. [7]

The sub-population formation algorithm is described as follows:
1. Population P of N members is divided into sub-populations SP1, SP2…SPf, where f is number of female members in P.
Each sup-population contains one female member.
2. For each male member in P do
Find the nearest female by calculating Euclidean Distance between male and each female
3. The male members are added to the sub-population related to the nearest female.

The problem with GAS3 algorithm is it does not guarantee that the females selected for species formation are uniformly distributed over the population. Initially in species formation process one female member along with male members are club together to form cluster. Later in merging of clusters into single cluster the female of dominated cluster is the female for merged cluster. In both these cases the cluster female may lay in near proximity of the boundary of a cluster. Such situation indicates that female members are not uniformly distributed over the population.

In GAS3KM, K-means clustering algorithm is applied on each cluster to move female member at the centre of cluster. Considering all the pros and cons of K-means, it is used after species formation phase. K-means require initial cluster centres as input. All the females are considered as initial cluster centres. To overcome problem associated with random choice of initial cluster centres, fertile females are selected as initial cluster centres. If data set is large then it is very difficult to guess proper number of clusters a priori. This problem also tackled by providing number of females as number of clusters.

It was our intuition that if we get well distributed and well separated clusters, it will enhance performance of GAS3 algorithm. So we have modified GAS3 algorithm. The modified pseudo-code is as follows
1. Initialize population

2. Use sex determination method to determine the sex of an individual in population
3. Species evolution phase creates many species around niches
4. Apply self-adaptive K-means algorithm.
5. Repeat
{
    5.1 For certain number of evolutions with each cluster in population
    do
    {
    5.1.1 *Selection Plan*: Choose one female parent and $\mu$ male parents from species S (the set *M*).
    5.1.2 *Generation Plan*: Create the offspring set *C* from *M*.
    5.1.3 *Replacement Plan*: Combine solutions in *C* and *M* to form set *R*. Arrange members in *R* according to their fitness.
    5.1.4 *Update Plan*: if one of the offspring is better than the female parent then replace that female parent with offspring. Otherwise compare offspring with male members and use replace worst strategy for updating.
    5.1.5 Increment the performance count of species if offspring replaces female parent.
    }
    5.2 Examine the performance of each species and merge species.
} Until (termination condition)

We call this new algorithm as GAS3KM (Modified Genetic Algorithm with Species and Sexual Selection using K-means).

## 5. Performance Comparison of GAS3KM with GAS3

Self-adaptive K-means algorithm is used in GAS3 algorithm for reclustering of population. This reclustering is needed for proper distribution of clusters in the search space. Well-distributed and well-separated clusters are required for efficient evolution of subpopulations.

    Our aim is to enhance performance of GAS3 algorithm. Since we are comparing our algorithm with GAS3, it becomes mandatory to use the same setup as used in [7]. The experimental setup used is given in Table 1. It uses SDM and species formation takes place around females. Drawback of GAS3 is that initial female selection does not guarantee that they are well distributed or well separated. GAS3KM uses K-means algorithm for further clustering. Four plans of co evolution are applied on these clusters. For the experimentation, fertile female scheme is used for SDM. Different values of R test effect of selection pressure on performance of the algorithm. Explorative recombination operator MLX (with $\eta= 4$) is used with five parents ($\mu=5$) to generate two offspring. MLX explores wide range of search space in the beginning. Proximity based species evolution scheme has constructed, the species around female parents. The parents, selected by sexual selection scheme undergo recombination or mutation operation. The exploitative

**Table 1.** GA Setup Used For Experimentation

| Parameter | Values |
|---|---|
| Population size (N) | 50,75,100,150,200 |
| Crossover probability parameter ($p_c$) | 0.3 - 0.8 in step of 0.1 |
| R (Related to selection pressure) | 1 - 10 in step of 1 |
| Stopping criteria | Maximum $10^6$ function evaluations or an objective value of $10^{-10}$ |
| Results average over | 50 independent runs |
| Parameters for performance evaluation | 1. Number of function evaluation for best run and worst run<br>2. Average number of function evaluation (AFES)<br>3. Best fitness, Average fitness and Worst fitness<br>4. Number of runs converged to global minima |
| Initialization of variables | Skewed initialization within [-10, -5] |
| Number of children ($\lambda$) | 2 |

recombination operator MPX (with $\eta= 1$) is used with the male parents selected randomly from species ($\mu<5$) and merging is performed after (N*N/R) number of evolution. Any good search algorithm must explore a large search space in the beginning and the search should then narrow down as it converges to the solution. To support this property explorative operator MLX is used in SDM and exploitative operator MPX is used in further evolution. 50 independent runs are executed for all possible combination of parameters R, N and $p_c$. Other parameters in GA are set as given in table 1.

### 5.1 Test functions

The minimization experiments will be performed on unconstrained unimodal and multi-modal functions with or without epitasis among n-variables as shown in table 2. Using skewed initialization these functions will be evaluated for global minima at 0.

### 5.2 Comparison of GAS3KM with GAS3

During experimentation it is found that initial positions of females changes after application of K-means algorithm. Females move away from each other and they spread well in the search space. Fig 1 shows initial positions of female and Fig 2 shows female positions after convergence of K-means algorithm. The reason behind this is that, in every iteration of K-means, females are updated with new mean value (centroid). unimodal and multimodal test function show that it has outperformed GAS3. It is observed that, GAS3KM

**Table 2** Definition of each test function together with its features

| Function | $f$ | Multimodal | Separable | Regular |
|---|---|---|---|---|
| Sphere | $f_{Elp} = \sum_{i=1}^{n} x_i^2$ | N | Y | n/a |
| Tablet | $f_{Tablet} = 10^6 x_1^2 + \sum_{i=2}^{n} x_i^2$ | N | Y | n/a |
| Cigar | $f_{Cigar} = x_1^2 + \sum_{i=2}^{n} 10^6 x_i^2$ | N | Y | n/a |
| Two-Axes | $f_{Two\_ax} = \sum_{i=1}^{[n/2]} 10^6 x_i^2 + \sum_{i=[n/2]}^{n} x_i^2$ | N | Y | n/a |
| Schwefel | $f_{Sch}(x) = \sum_{i=1}^{n} (\sum_{j=1}^{i} x_j)^2$ | N | N | n/a |
| Rosenbrock | $f_{Rosen}(x) = \sum_{i=1}^{n-1} (100(x_i^2 - x_{i+1})^2 + (x_i - 1)^2)$ | Y | N | n/a |
| Rastrigin | $f_{Rst}(x) = 10n + \sum_{i=1}^{n} (x_i^2 - 10\cos(2\pi x_i))$ | Y | Y | n/a |
| Scaled Rastrigin | $f_{rastScaled}(x) = 10n + \sum_{i=1}^{n} ((x_i 10^{\frac{i-1}{n-1}})^2 - 10\cos(2\pi x_i))$ | Y | Y | n/a |
| Skew Rastrigin | $f_{rastSkew}(x) = 10n + \sum_{i=1}^{n} (y_i^2 - 10\cos(2\pi x_i))$ with $y_i = \begin{cases} 10.x_i & if\ x_i > 0, \\ x_i & otherwise \end{cases}$ | Y | Y | n/a |
| Griewangk | $f_{Grie} = \frac{1}{4000} \sum_{i=1}^{n} x_i^2 - \prod_{i=1}^{n} \cos\left(\frac{x_i}{\sqrt{i}}\right) + 1$ | Y | N | Y |
| Ackley | $f_{Ack}(x) = -20 \cdot \exp\left(-0.2 \cdot \sqrt{\frac{1}{n} \sum_{i=1}^{n} x_i^2}\right) - \exp\left(\frac{1}{n} \sum_{i=1}^{n} \cos(2.0 * 3.14 * x_i)\right) + 20 + e^1$ | Y | N | Y |
| Bohachevsky | $f_{Bohachevaky}(x) = \sum_{i=1}^{n-1} (x_i^2 + 2x_{i+1}^2 - 0.3\cos(3\pi x_i) - 0.4\cos(4\pi x_{i+1}) + 0.7)$ | Y | N | n/a |

As a consequence of this, females change their positions. Actually female selection strategy used in GAS3 does not guarantee that selected females are well distributed in the search space and hence does not form well distributed and well separated clusters. It was our intuition that if we get well distributed and well separated clusters, it will enhance performance of GAS3 algorithm.

**5.2.1 Performance Test under different selection pressure(R):** Experimental results of GAS3KM with requires less number of function evaluations for finding minima. Its behaviour is similar to GAS3 for unimodal functions as the value of R increases; number of function evaluations (FES) decreases, and for multi-modal functions as the value of R increases, and FES also increases. GAS3KM has performed better, with unimodal functions for high value of R and with multi-modal function for low value of R. Since the parameter R plays important role in SDM and merging of species of the algorithm. The low value of R means the population members gets more chance to prove its fertility to become a female member. This creates a high selection pressure that results into well-defined niches in population to solve multi-modal functions efficiently. Also the low value of R lengthens the duration (i.e. number of evolutions) to perform the merging, helps the species to prove their survival by exploring/exploiting search space. The high value of R sets a low selection pressure and narrow down the duration of merging that helps in solving unimodal functions. Figure 3, 4 and 5 shows performance of GAS3KM on unimodal functions against R (with $p$c =0.5 and N=100). For ellipsoidal function GAS3KM shows approximately 50% reduction in average number of function evaluations. For the Schwefel function GAS3KM has performed well with moderated value of R. For the Rosenbrock function it has not given proper performance as shown in figure 5. Figures 6,7 and 8 shows performance of the GAS3KM on multi-modal functions against the parameter R (with $p$c =0.3 and N=100).

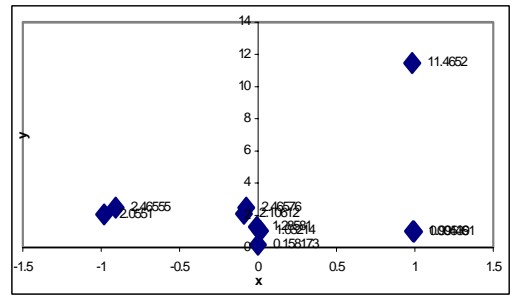

**Figure 1.** Initial positions of females

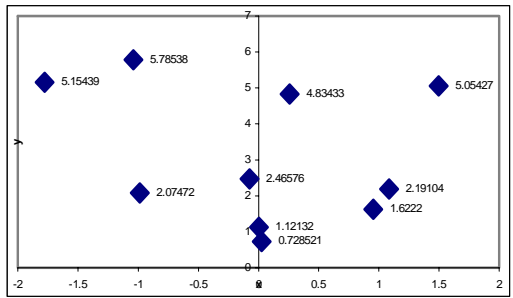

**Figure 2.** Female positions after K-means

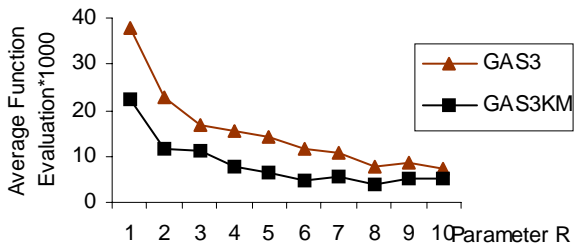

**Figure 3.** Performance Test on the Ellipsiodal Function

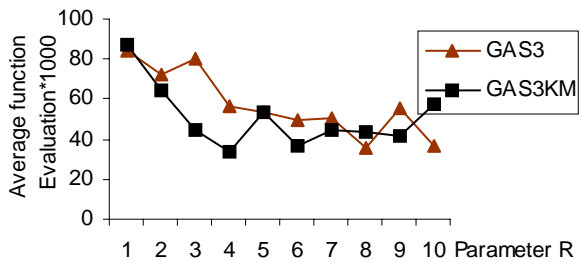

**Figure 4.** Performance Test on the Schwefel Function

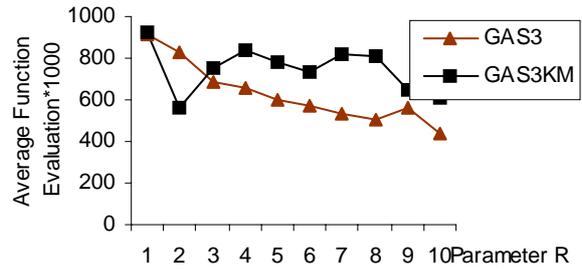

**Figure 5.** Performance Test on the Rosenbrock Function

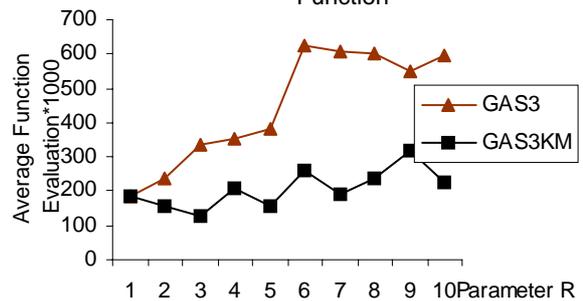

**Figure 6.** Performance Test on the Rastrigin Function

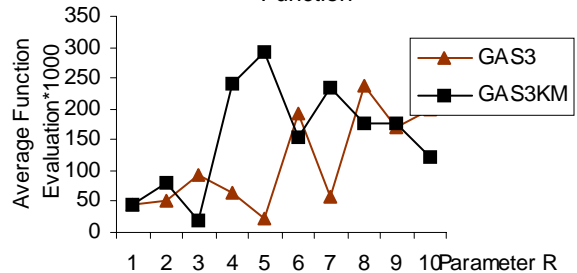

**Figure 7.** Performance Test on the Griewangk Function

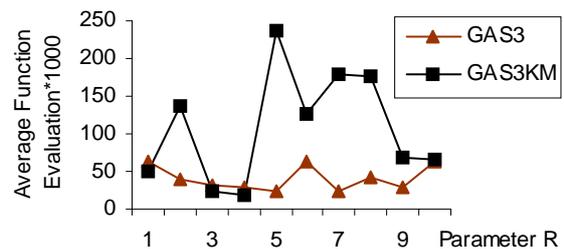

**Figure 8.** Performance Test on Ackley Function

**Table 3** Performance of GAS3KM on Unimodal functions Against Population Size N

| N | Ellipsoidal | | | | Schwefel | | | | Rosenbrock | | | |
|---|---|---|---|---|---|---|---|---|---|---|---|---|
| | GAS3 | | **GAS3KM** | | GAS3 | | **GAS3KM** | | GAS3 | | **GAS3KM** | |
| | AFES | Success | AFES | Success | AFES | Success | AFES | Success | AFES | Success | AFES | Succes |
| 50 | **5158.13** | 100% | **3959.56** | 100% | 44060.9 | 100% | 36328.6 | 100% | 401162 | 86.67% | 691402 | 62.00% |
| 75 | 7101.64 | 100% | 6240.17 | 100% | 55176.6 | 100% | **28326.8** | **100%** | 541917 | 66.67% | 705763 | 68.00% |
| 100 | 7331.84 | 100% | 7363.39 | 100% | **36570.7** | **100%** | 57777.1 | 100% | **434632** | **93.33%** | 585133 | 77.78% |
| 150 | 15853.6 | 100% | 9379.39 | 100% | 58657.4 | 100% | 37035.7 | 100% | 608751 | 91.11% | 729207 | 77.78% |
| 200 | 25023.3 | 100% | 16356.7 | 100% | 82566.2 | 100% | 84106.1 | 100% | 856712 | 48.89% | 835245 | 41.12% |

**Table 4** Performance of GAS3KM on Multimodal functions Against Population Size N

| | Rastrigin | | | | Griewangk | | | | Ackley | | | |
|---|---|---|---|---|---|---|---|---|---|---|---|---|
| | GAS3 | | **GAS3KM** | | GAS3 | | **GAS3KM** | | GAS3 | | **GAS3KM** | |
| N | AFES | Success | AFES | Succes | AFES | Success | AFES | Success | AFES | Success | AFES | Succes |
| 50 | 374398 | 71.11% | **97787.4** | **100%** | 38604.7 | 97.78% | **15613.7** | **100%** | **22308.9** | **100%** | 73287.5 | 95.22% |
| 75 | 229172 | 91.11% | 143810 | 100% | **30346.1** | **100%** | 18500.8 | 100% | 41447.2 | 100% | **29561.9** | **100%** |
| 100 | **187978** | **100%** | 135082 | 100% | 46323.2 | 100% | 45476.6 | 100% | 62702.8 | 100% | 50945.6 | 100% |
| 150 | 224996 | 100% | 269801 | 100% | 87217 | 100% | 85831.1 | 100% | 118166 | 100% | 101091 | 100% |
| 200 | 322114 | 100% | 316258 | 100% | 138785 | 100% | 139117 | 100% | 183287 | 100% | 139117 | 100% |

**5.2.2 Effect of population size:** GAS3KM is tested for different size of population. For all population sizes it has given better performance than GAS3. Table 3 shows performance comparison of GAS3KM with GAS3 for unimodal test functions against population size N. Table 4 shows performance comparison of GAS3KM with GAS3 for multimodal test functions against population size N. For different population size it has shown relative increase in number of function evaluation with the population size.

Table 5 shows performance of GAS3KM on Unimodal and Multimodal functions. GAS3KM has achieved almost 100% success in solving all unimodal test functions. For the Rosenbrock function it has not given proper performance. Rosenbrock's valley is also known as Banana function. The global optimum is inside a long, narrow, parabolic shaped flat valley and to find the valley is trivial, however convergence to the global optimum is difficult. For all multimodal test function it has given 100% success. From the above Table it is clear that there is significant change in number of function evaluations required to find minima.

## 6. Conclusion

GAS3KM successfully implemented and tested on all unimodal and multimodal test function against parameters R and N. GAS3KM has combined simplicity of K-means and robust nature of GA.

The test results of GAS3KM for unimodal and multimodal test function show that it outperforms GAS3 algorithm in terms of number of function evaluation. K-means algorithm helps in producing well-distributed and well-separated clusters. These clusters evolve more quickly and require less number of function evaluations for finding optima. K-means also inject diversity in female population. K-means used with Sex Determination Method and Species Evolution Phase, behaves like self-adaptive algorithm, which automatically detects number of clusters and initial position of centroids.

Thus, GAS3KM is a robust genetic algorithm for efficient evolution of distributed population. GAS3KM works for single-objective optimization. GAS3KM has shown positive response to diversity preservation and controls premature convergence of GA. In this work we have concentrated on female selection and cluster distribution and separation. Future study will be on strategies for selection of parameters like number of clusters and effect of size of clusters on convergence of algorithm. Further we will study new plans for selection of males for crossover and updated GAS3KM for multi-objective optimization.

Table 5 Test result of GAS3km On Unimodal Function and Multimodal Function

| Function | GAS3/GAS3km | No. Of Function Evaluations | | | Fitness | | | Success |
|---|---|---|---|---|---|---|---|---|
| | | Best Run | Avg Run | WrstRun | Best | Average | Worst | |
| **Test result of GAS3km On Unimodal Function** | | | | | | | | |
| Ellipsoidal | GAS3 | 4071 | 7331.84 | 19546 | 2.57E-11 | 7.43E-11 | 9.97E-11 | 100.00% |
| | GAS3km | 5085 | 7363.39 | 14554 | 6.59E-11 | 8.88E-11 | 9.96E-11 | 100.00% |
| Schwefel | GAS3 | 17029 | 36570.7 | 137041 | 4.68E-11 | 8.62E-11 | 9.97E-11 | 100.00% |
| | GAS3km | 7000 | 57777.1 | 145940 | 4.50E-11 | 8.77E-11 | 9.95E-11 | 100.00% |
| Sphere | GAS3 | 3427 | 7786.53 | 16154 | 4.53E-11 | 8.84E-11 | 9.98E-11 | 100.00% |
| | GAS3km | 3367 | 7449.5 | 15726 | 5.70E-11 | 8.75E-11 | 9.99E-11 | 100.00% |
| Cigar | GAS3 | 3758 | 11630.8 | 25248 | 5.15E-11 | 8.46E-11 | 9.96E-11 | 100.00% |
| | GAS3km | 3965 | 14282.4 | 24824 | 7.13E-11 | 9.11E-11 | 9.99E-11 | 100.00% |
| Tablet | GAS3 | 4305 | 13251.4 | 26751 | 6.73E-11 | 8.75E-11 | 9.86E-11 | 100.00% |
| | GAS3km | 3479 | 5600.7 | 12226 | 6.04E-11 | 8.75E-11 | 9.99E-11 | 100.00% |
| TwoAxed | GAS3 | 7735 | 14523.9 | 46451 | 4.94E-11 | 8.76E-11 | 9.93E-11 | 100.00% |
| | GAS3km | 4151 | 13384.5 | 37934 | 5.63E-11 | 8.65E-11 | 9.86E-11 | 100.00% |
| Rosenbrock | GAS3 | 195748 | 434632 | 1000002 | 9.90E-11 | 8.86E-02 | 3.99E+00 | 93.33% |
| | GAS3km | 194218 | 585133 | 1000002 | 9.43E-11 | 0.443151 | 3.98666 | 77.78% |
| **Test result of GAS3km On Multimodal Function** | | | | | | | | |
| Rastrigin | GAS3 | 68125 | 187978 | 412180 | 1.18E-11 | 7.48E-11 | 9.98E-11 | 100.00% |
| | GAS3km | 91380 | 135082 | 182043 | 1.52E-11 | 7.64E-11 | 9.93E-11 | 100.00% |
| Rastrigin Scaled | GAS3 | 59384 | 145779 | 294441 | 5.91E-11 | 6.72E-11 | 9.98E-11 | 100.00% |
| | GAS3km | 63322 | 156998 | 323350 | 1.18E-11 | 6.32E-11 | 9.94E-11 | 100.00% |
| Rastrigin Skewed | GAS3 | 62895 | 216585 | 1000002 | 4.43E-11 | 0.04422 | 0.99495 | 95.56% |
| | GAS3km | 76140 | 196754 | 602895 | 1.97E-11 | 6.99E-11 | 9.46E-11 | 100.00% |
| Ackley | GAS3 | 54738 | 62702 | 70121 | 5.92E-11 | 8.96E-11 | 9.99E-11 | 100.00% |
| | GAS3km | 37489 | 50945.6 | 69028 | 7.43E-11 | 9.31E-11 | 9.88E-11 | 100.00% |
| Griewangk | GAS3 | 40376 | 46323.2 | 49366 | 3.78E-11 | 8.30E-11 | 1.00E-10 | 100.00% |
| | GAS3km | 36460 | 45476.6 | 54503 | 7.57E-11 | 8.55E-11 | 9.67E-11 | 100.00% |
| Bohachevsky | GAS3 | 42828 | 50335.9 | 54937 | 3.85E-11 | 8.28E-11 | 9.99E-11 | 100.00% |
| | GAS3km | 30809 | 43562.8 | 48715 | 6.53E-11 | 8.41E-11 | 9.99E-11 | 100.00% |